\pdfoutput=1
\documentclass[journal]{IEEEtran}

\ifCLASSINFOpdf
\else
   \usepackage[dvips]{graphicx}
\fi
\usepackage{url}

\usepackage{graphicx}

\usepackage{multirow}
\usepackage[table,xcdraw]{xcolor}
\usepackage{booktabs}
\usepackage{makecell}
\usepackage{subfigure}  %插入多图时用子图显示的宏包
\usepackage{bm}
\usepackage{amsfonts} 
\usepackage{balance}
\usepackage{cite}

\begin{document}

\title{Breaking Free from Fusion Rule: A Fully Semantic-driven Infrared and Visible Image Fusion}

\author{Yuhui Wu, Zhu Liu, Jinyuan Liu, Xin Fan, \IEEEmembership{Senior Member, IEEE}, Risheng Liu, \IEEEmembership{Member, IEEE}

\thanks{This work is partially supported by the National Key R\&D Program of China (2020YFB1313503), the National Natural Science Foundation of China (Nos. 61922019 and U22B2052),  the Fundamental Research Funds for the Central Universities and the Major Key Project of PCL (PCL2021A12). (\textit{Corresponding author: Risheng Liu}.)
	
Yuhui Wu and Zhu Liu are with the School of Software Technology, Dalian University of Technology, Dalian 116024, China. (wuyuhui63@hotmail.com; liuzhu\_dlut@mail.dlut.edu.cn).

Jinyuan Liu is with the School of Mechianical of Engineering, Dalian University of Technology, Dalian 116024, China. (e-mail: atlantis918@hotmail.com).

Xin Fan and Risheng Liu are with the DUT-RU International School of Information Science \& Engineering, Dalian University of Technology, Dalian 116024, China. They are also with Peng Cheng Laboratory, Shenzhen 518052, China. (e-mail: xin.fan@dlut.edu.cn; rsliu@dlut.edu.cn).}}

\markboth{Journal of \LaTeX\ Class Files, Vol. 14, No. 8, August 2015}
{Shell \MakeLowercase{\textit{et al.}}: Bare Demo of IEEEtran.cls for IEEE Journals}
\maketitle

\begin{abstract}
Infrared and visible image fusion plays a vital role in the field of computer vision. Previous approaches make efforts to design various fusion rules in the loss functions. However, these experimental designed fusion rules make the methods more and more complex. Besides, most of them only focus on boosting the visual effects,  thus showing unsatisfactory performance for the follow-up high-level vision tasks. To address these challenges, in this letter, we develop a semantic-level fusion network to sufficiently utilize the semantic guidance, emancipating the experimental designed fusion rules. In addition, to achieve a better semantic understanding of the feature fusion process, a fusion block based on the transformer is presented in a multi-scale manner. Moreover, we devise a regularization loss function, together with a training strategy, to fully use semantic guidance from the high-level vision tasks. Compared with state-of-the-art methods, our method does not depend on the hand-crafted fusion loss function. Still, it achieves superior performance on visual quality along with the follow-up high-level vision tasks.
\end{abstract}


\IEEEpeerreviewmaketitle

\section{Introduction}
\IEEEPARstart{I}{nfrared} and visible image fusion (IVIF), referring to providing the typical characteristics from source images, has been witnessed rapid development in recent years. IVIF can effectively break the limitation of information loss from single sensor and plays an important role
for the follow-up high-level vision tasks, e.g., object detection~\cite{person}, semantic segmentation~\cite{MFNet} and so on. Unfortunately, there are few works~\cite{TarDAL,SeAFusion,DetFusion} to bridge high-level semantic tasks with image fusion. In this paper, rather than explicitly construct fusion rules for IVIF, we propose a generic semantic-driven learning  paradigm to investigating  task-specific image fusion.

%\IEEEPARstart{I}{nfrared} and visible image fusion (IVIF) has a wide range of applications in video surveillance \cite{Surveillance}, person detection \cite{person}, and semantic segmentation \cite{MFNet}, etc. This is because a single modality sensor can only capture limited information. Concretely, visible light imaging cameras can capture accurate color information like human vision. But harsh atmospheric conditions can disturb visible light imaging, e.g., haze, fog, and low or bright light. Infrared images capture thermal radiation at the site, showing attention-grabbing pedestrians, animals, and moving cars, but lack sufficient texture information to describe the scene. Therefore, the complementarity of these two imaging mechanisms motivates us to fuse the infrared and visible images into a single image. 

In the past, traditional fusion methods prevailed, including \cite{SR2, MST1, GTF, hara2014differentiable, subspace1, 8627371, 7906563, 8506435}. Recently, deep learning-based methods have shown great potential in numerous image-processing fields \cite{ jiang2022towards, 8352542, Ma_2022_CVPR} by their powerful nonlinear feature extraction capabilities. Learning-based IVIF approaches achieve promising performance based on the design of various fusion rules and loss functions. We can roughly divide current learning schemes into two categories: fusion-rule based methods~\cite{DenseFuse,RFN-Nest,SMoA, vs2022image} and end-to-end learning schemes~\cite{TarDAL, U2Fusion, STDFusion, SwinFusion, FusionGAN, SF-Net, SeachHierarchical, liu2022learn, wang2022unsupervised, zhang2020rethinking, liu2022attention}.

 In specific, the first kind of fusion methods rely on the manually designed rules to aggregate the modal feature approximately. These methods firstly utilize the auto encoder mechanisms to extract and reconstruct multimodal features to sufficiently learn the significant feature extraction. Then they develop various fusion rules for feature fusion, e.g., weighted average, summation, maximum selection, and $\ell_{1}$ norm. For instance, Li \textit{et al.}~\cite{DenseFuse} pioneered the dense blocks as the learnable auto-encoder and utilized the weighted-average strategy to fuse modal features. After that, Li \textit{et al.}~\cite{RFN-Nest} also provide spatial/channel attention mechanisms as the fusion strategies to fuse features with nested connections. Subsequently, Liu \textit{et al.}~\cite{MFEIF} introduces the edge attention guided auto-encoder to extract feature and adopts the simple fusion rules. 
%Subsequently, Liu \textit{et al.} propose the architecture search-based scheme to discover modal-oriented network and introduce self-visual saliency strategy to promote the visual quality.

We can obviously observe that,  current methods rely on the appropriate fusion strategies to guide the feature fusion. However, these fusion strategies are not sensitive to the diverse data distribution and are easy to induce visual artifacts and blurs. More importantly, the manual design of fusion strategies are too fragile to preserve suitable modal characteristics for supporting follow-up high-level vision tasks.

Instead of manually designed fusion rules, end-to-end learning methods are proposed to establish the connection between source and fused images directly. Specifically, architectures and loss functions are two challenging stumbling stones for theses methods. Existing methods concentrate on designing architectures based on the current effective practices, rather than considering the particular properties of fusion tasks. For example, dense blocks~\cite{U2Fusion} and residual blocks~\cite{SeAFusion} are widely utilized for IVIF. 
Besides, diverse loss functions are proposed to enforce different principles for IVIF. Typically,  Ma \textit{et al.}~\cite{FusionGAN} introduce the dual generative adversarial criterion to push the generated images as similar as the source images. Xu \textit{et al.}~\cite{U2Fusion} propose the feature-level
measurement to endow the richness of information into fused images. Tang \textit{et al.}\cite{SeAFusion} uses the max-choosing rule to design loss function, but is limited in extreme scenes (e.g., glare and fog). 

Though visually appealing results which can acquire remarkable statistical metrics are obtained. These loss functions coupled with training strategies make the methods more complex. Furthermore, these architectures cannot effectively extract  the modal characteristics, limited by the local awareness of  convolution networks. We argue that both two categories of methods are designed to improve the visual quality of fusion, neglecting the requirement of follow-up semantic tasks.

To partially mitigate these issues, in this letter, we propose a  semantic-driven fusion method. Instead of considering the image fusion as an independent task, we fully leverage the guidance of high-level semantic tasks to reserve the beneficial information and reduce the conflicts. In this way, our fused results not only highlight the comprehensive information, but also facilitate the following semantic tasks. To be concrete, we first propose a multi-scale fusion network with self-attention mechanism to sufficiently aggregate the  modal features. Multi-scale  extraction can effectively combine features in a coarse-to-fine manner from the scene structure to context details. Self-attention mechanism is to establish the long-range dependency of multi-modal features, better depict the global 
representation of salient targets. Then we introduce a correlated regularization to describe the relationships between source images and fused images. Based on this, we only utilize the criterion of high-level vision tasks to train both the fusion and high-level networks. Thus, this strategy emancipates the experimental design of fusion rules, discards the restriction of modal statistic metrics and drastically improves the performance of high-level vision tasks.
 We summarize the core contribution as follows:
\begin{itemize}
	\item A multi-scale self-attention-based image fusion network is proposed to effectively represent the global structures in a coarse-to-fine manner.
	\item Imposing a correlated regularization, a fully semantic-driven training strategy is introduced to break free from handcrafted fusion rules.
\end{itemize}
%Meanwhile, to pursue high efficiency, deep learning-based image fusion methods greatly simplified the fusion network because few convolution operations are capable to fit the handcrafted fusion loss function. However, such a simplified network cannot accomplish semantic-level fusion. In other words, the fusion network should understand the semantic information of source images and flexibly fuse them.
%
%To address these issues, in this letter, we propose a semantic-driven training strategy. Specifically, we pre-train the fusion network for a basic fusion ability during the warm-start phase. And then during the semantic training phase, we use the semantic segmentation task to fine-tune the fusion network. This strategy can solve the distortion problem of fusion task, as shown in Fig.~\ref{fig1d}. On the other hand, to accomplish semantic-level fusion, we develop a global visual perception-oriented fusion network aiming to make full use of guidance from the high-level vision tasks.
%The main contributions are:
%\begin{itemize}
%	\item We propose a global visual perception-oriented fusion network to sufficiently utilize guidance from high-level vision tasks and provide the semantic-level fusion for high-level vision tasks.
%	\item More importantly, we develop a semantic-driven training strategy to flexibly adjust the proportion of infrared and visible information using the segmentation task, avoiding designing rigid fusion loss functions.
%	\item Extensive experiments demonstrate the superiority of our method compared to state-of-the-art methods.
%\end{itemize}

\section{The Proposed Method}

\subsection{Network Architecture}
While some previous works focus on pursuing high efficiency at the expense of the fitting ability of the fusion network, it does not work well to utilize the semantic guidance from the follow-up high-level vision tasks. In contrast, we hypothesize that the fusion network should be capable of semantic understanding to achieve flexible fusion effects among various semantic classes.

\begin{figure}[htbp]
	
	\centering
	\begin{tabular}{c}
		\includegraphics[width=0.95\linewidth]{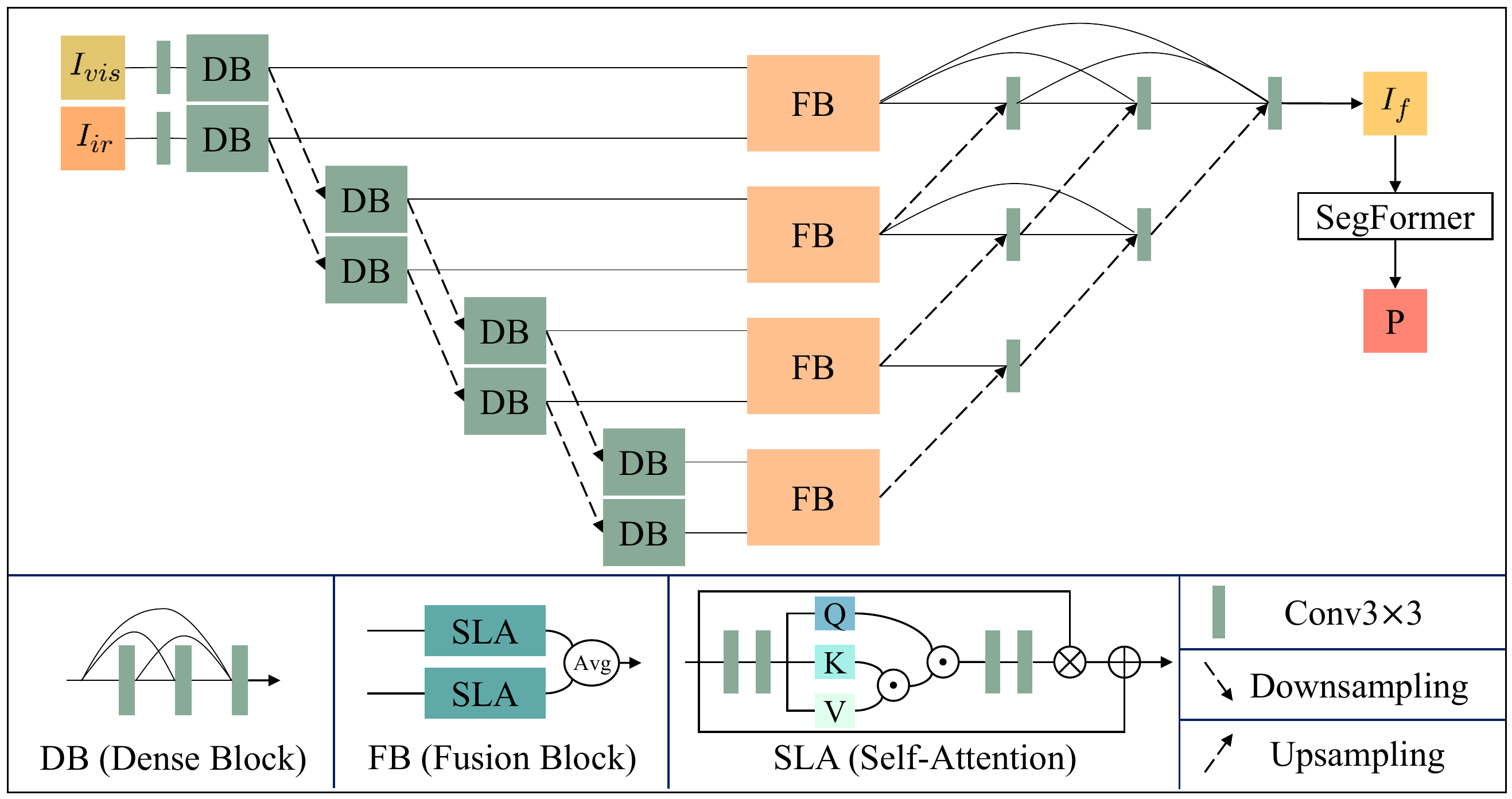}
	\end{tabular}
	\caption{The workflow of the proposed method. $Avg$, $\otimes$ and $\oplus$ represent the computation of element-wise average computation, multiplication, and addition, respectively. $\odot$ means matrix multiplication. P denotes the  prediction.}
	\label{Network}
\end{figure}

To this end, we employ the multi-scale mechanism from \cite{RFN-Nest} to deal with the textural details and semantic information, respectively. As shown in Fig.~\ref{Network}, we use the downsampling (i.e. max-pooling) operations to obtain feature maps of different resolutions.
 Among them, the shallow layer feature maps contain more textures information, while the deep layer feature maps contain more semantic information. 
 
% The multi-scale mechanism provides semantic understanding ability for the fusion network.

On the other hand, to fuse the extracted cross-modal feature maps, we devise a generic fusion block based on the efficient self-attention \cite{EfficientAttention}. As shown in the bottom of Fig.~\ref{Network}, the fusion block consists of two self-attention modules, which can capture and reinforce the useful components in the global receptive field. In the self-attention module, we reshape the feature maps of $\mathbb{R}^{H\times W \times C}$ to vectors of $\mathbb{R}^{N \times C}$, where $ N = H\times W $. Then, we use the linear layers to encode the vectors into quary Q, key K, and value V. We get the attention map by a matrix multiplication $K^{T}V$, then we acquire the final attention result $Q\mbox{Softmax}(K^{T}V)$. The strengthened component is obtained by element-wise multiplying the attention result and the input feature maps. To reserve the detail information, we further introduce a residual connection.

%\begin{figure}[htb]
%	
%	\centering
%	\begin{tabular}{c}
%		\includegraphics[width=0.95\linewidth]{results/Pipeline.pdf}
%	\end{tabular}
%	\caption{The proposed semantic-driven training strategy. }
%	\label{Pipeline}
%\end{figure}

\vspace{-0.4cm} 
\subsection{Training Strategy}
	Existing end-to-end deep-learning methods focus on devising fusion rules to acquire visually appealing results. Unfortunately, hand-crafted fusion rules are heavily limited to the scene and cannot meet the essential requirement of follow-up semantic tasks. To address this issue, we develop a semantic-driven training strategy to emancipate the manual design. %, as shown in Fig.~\ref{Pipeline}.
\subsubsection{Warm-start phase}
	Jointly training both the fusion and segmentation networks is an intuitive strategy. However, at the beginning of training, the parameters of the fusion model are randomly initialized and thus cannot provide meaningful fused images for the segmentation network to handle. Consequently, the training process deviates from our expectations.
	
%	 Introducing a hand-crafted fusion loss function is an intuitive solution. However, the hand-crafted fusion loss function cannot meet the requirement of subsequent high-level vision tasks. 
	
	To address this issue, we use an average strategy to pre-train the fusion model to obtain a malleable initialization.
	This learning procedures can be formulated as
	\begin{equation}
		\min\limits_{\bm{\theta}}\mathcal{L}_{WS}\left ( \mathcal{N}_{F}\left ( I_{vis}, I_{ir}; \bm{\theta} \right )  \right ),
	\end{equation}
	where $\mathcal{N}_{F}$ is the fusion network with learnable parameters $\bm{\theta}$, $I_{ir}$ and $I_{vis}$ denote infrared image and visible image, respectively. After this phase, we obtain $\bm{\theta^{'}}$, which can fuse the source images into substantially average and meaningful results for the next training procedure.
	
\subsubsection{Semantic training phase}
%	The $\bm{\theta^{'}}$ cannot meet the requirement of both human vision and the segmentation task.
	In this phase, we fine-tune the fusion network by jointly training with the segmentation network, which can be formulated as
	\begin{equation}
		\min\limits_{\bm{\theta},\bm{\omega}}\mathcal{L}_{ST}\left( \mathcal{N}_{S}\left ( \mathcal{N}_{F}\left ( I_{vis}, I_{ir}; \bm{\theta^{'}} \right ) ; \bm{\omega}\right) \right ) ,
	\end{equation}
	where the $\mathcal{N}_{S}$ is the semantic segmentation network with learnable parameters $\bm{\omega}$. 
	The semantic segmentation task will learn to adjust the proportion of infrared and visible components away from the average fusion state.
	
	It is instructive to note that the fusion model without additional constraints is unstable, resulting in degraded segmentation performance. To mitigate this, we design an auxiliary regularization loss function to constrain the fusion model for utilizing far-reaching semantic guidance. 

\begin{figure*}[htbp]
	\centering
	\setlength{\tabcolsep}{0.5pt}
	\footnotesize
	\begin{tabular}{ccccccccccc}
		\includegraphics[width=0.088\textwidth]{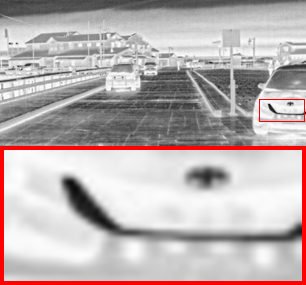}&
		\includegraphics[width=0.088\textwidth]{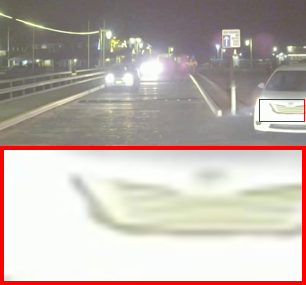}&
		\includegraphics[width=0.088\textwidth]{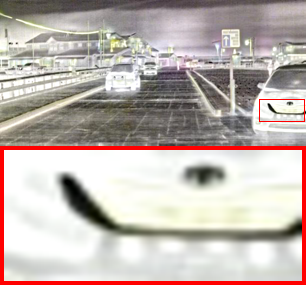}&
		\includegraphics[width=0.088\textwidth]{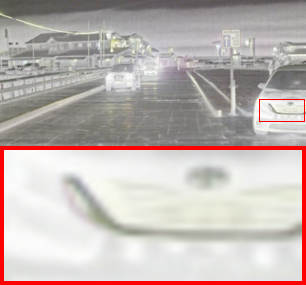}&
		\includegraphics[width=0.088\textwidth]{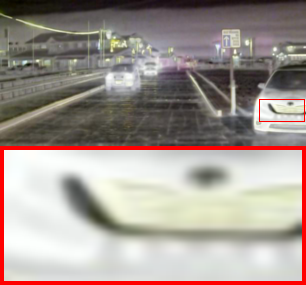}&
		\includegraphics[width=0.088\textwidth]{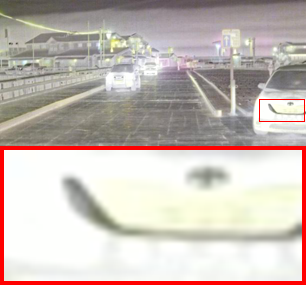}&
		\includegraphics[width=0.088\textwidth]{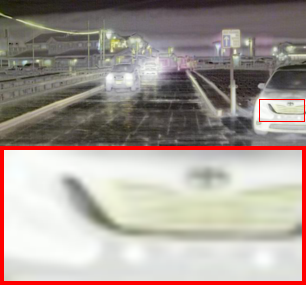}&
		\includegraphics[width=0.088\textwidth]{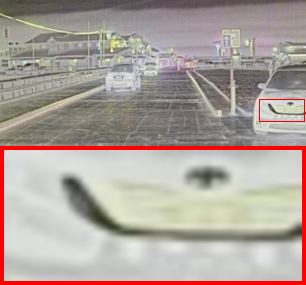}&
		\includegraphics[width=0.088\textwidth]{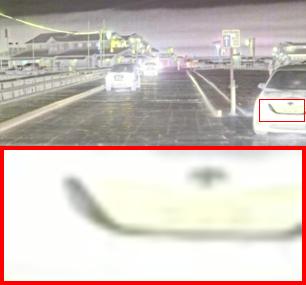}&
		\includegraphics[width=0.088\textwidth]{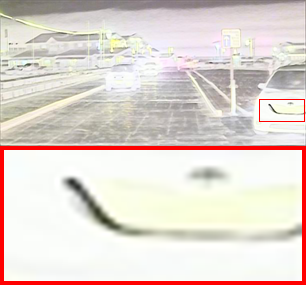}&
		\includegraphics[width=0.088\textwidth]{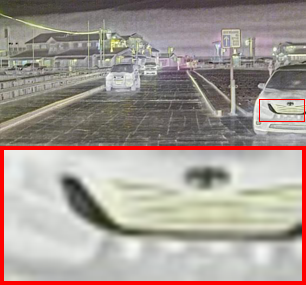}\\
		
		\includegraphics[width=0.088\textwidth]{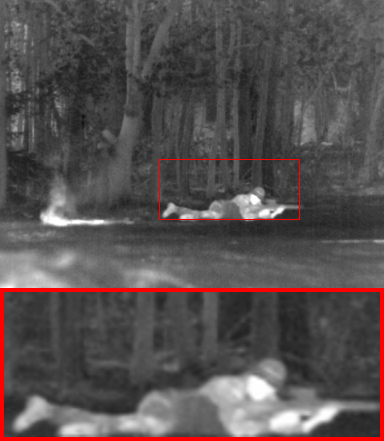}&
		\includegraphics[width=0.088\textwidth]{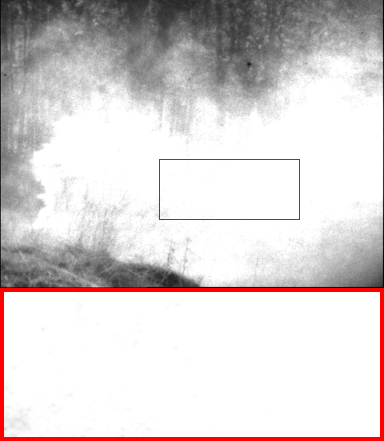}&
		\includegraphics[width=0.088\textwidth]{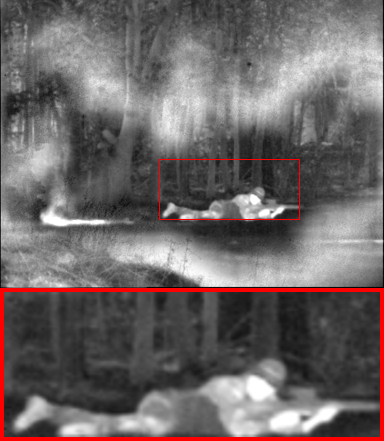}&
		\includegraphics[width=0.088\textwidth]{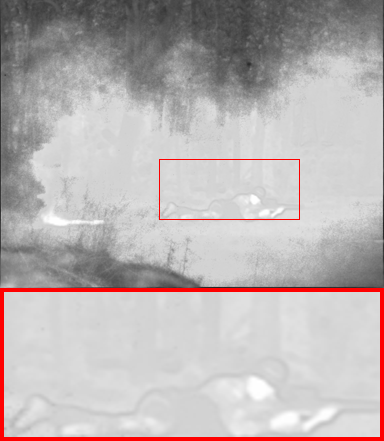}&
		\includegraphics[width=0.088\textwidth]{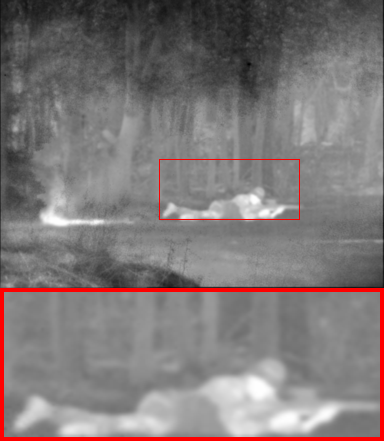}&
		\includegraphics[width=0.088\textwidth]{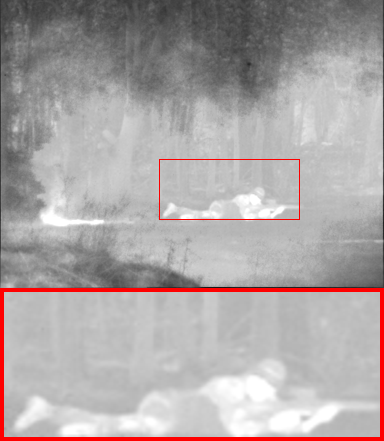}&
		\includegraphics[width=0.088\textwidth]{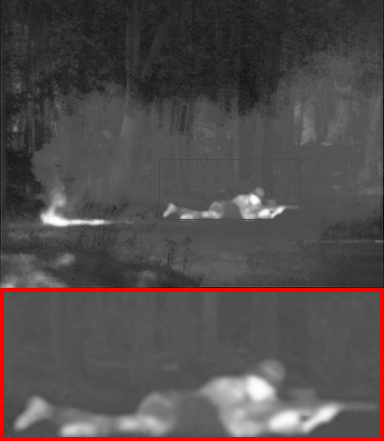}&
		\includegraphics[width=0.088\textwidth]{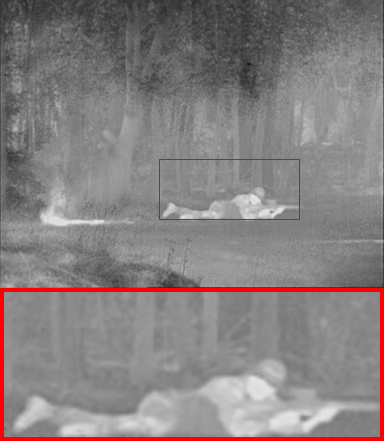}&
		\includegraphics[width=0.088\textwidth]{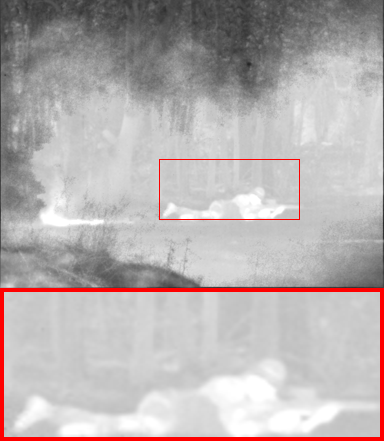}&
		\includegraphics[width=0.088\textwidth]{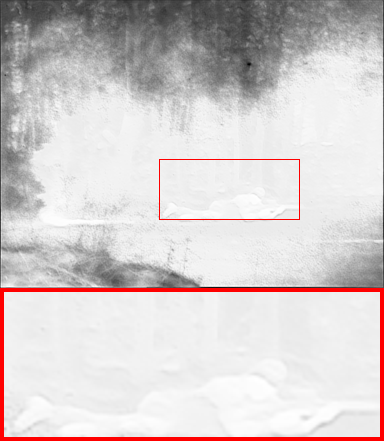}&
		\includegraphics[width=0.088\textwidth]{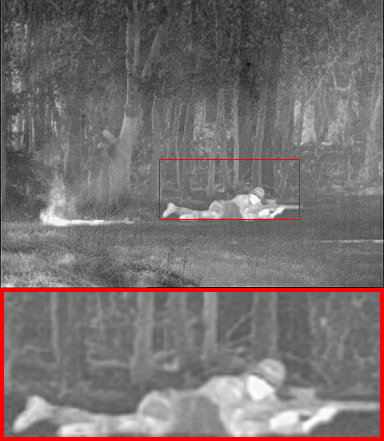}\\
		
		\includegraphics[width=0.088\textwidth]{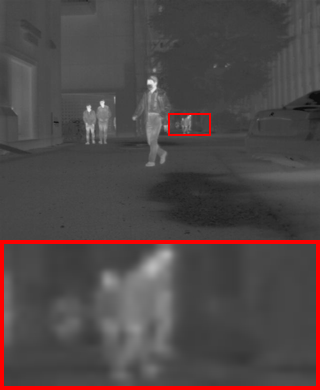}&
		\includegraphics[width=0.088\textwidth]{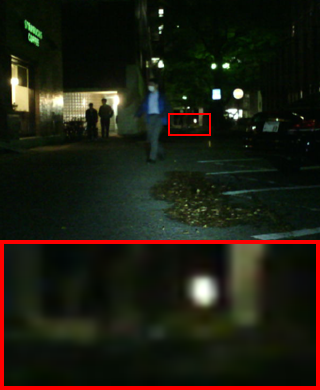}&
		\includegraphics[width=0.088\textwidth]{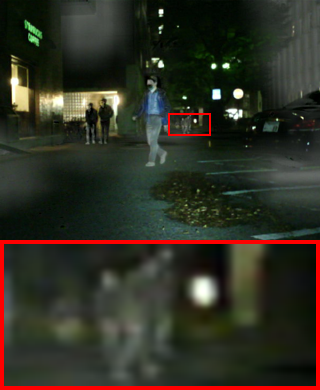}&
		\includegraphics[width=0.088\textwidth]{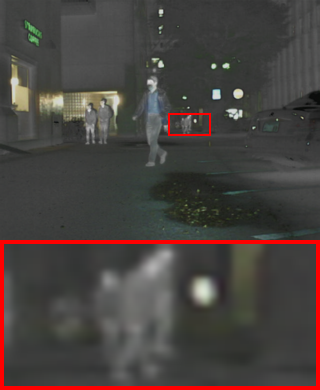}&
		\includegraphics[width=0.088\textwidth]{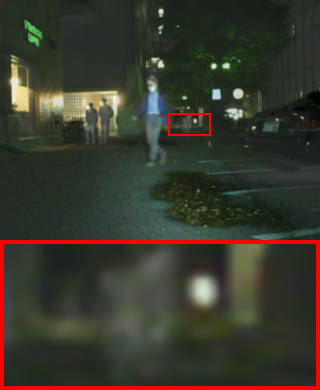}&
		\includegraphics[width=0.088\textwidth]{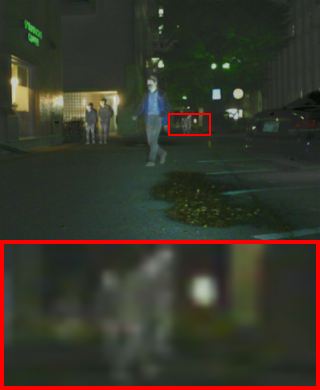}&
		\includegraphics[width=0.088\textwidth]{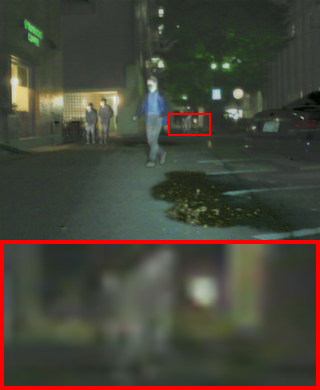}&
		\includegraphics[width=0.088\textwidth]{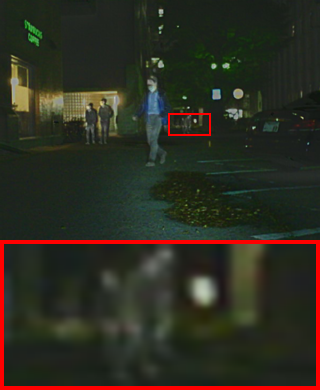}&
		\includegraphics[width=0.088\textwidth]{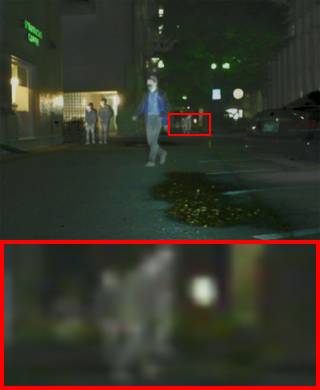}&
		\includegraphics[width=0.088\textwidth]{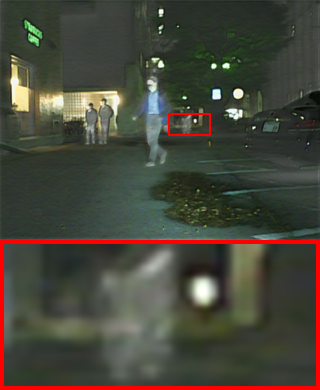}&
		\includegraphics[width=0.088\textwidth]{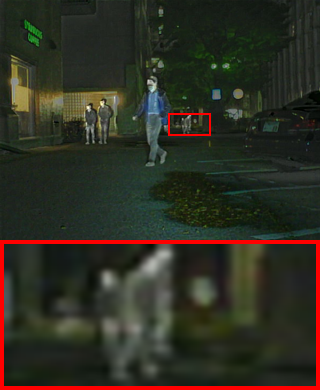}
		\\
		Ir&Vis&MST-SR&DenseFuse&RFN-Nest&SMoA&GANMcC&U2Fusion&MFEIF&SeAFusion&Ours
		
	\end{tabular}
	\caption{Qualitative comparison with state-of-the-art fusion methods. The three rows of results are from the RoadScene, TNO, and MFNet datasets, respectively.}
	\label{fig: Qualitative}
\end{figure*}

\begin{table*}[!htb]
	\centering
	\renewcommand\arraystretch{1.2} 
	\setlength{\tabcolsep}{1.9mm}
	\footnotesize
	\caption{Per-class segmentation results on the MFNet dataset. The best result is in {\textcolor{red}{\textbf{red}}} whereas the second best one is in {\textcolor{blue}{\textbf{blue}}}.}
	\begin{tabular}{|c|cc|cc|cc|cc|cc|cc|cc|cc|cc|cc|}
		\hline
		\multirow{2}{*}{Method}&\multicolumn{2}{c|}{Car}&\multicolumn{2}{c|}{Person}&\multicolumn{2}{c|}{Bike}&\multicolumn{2}{c|}{Curve}&\multicolumn{2}{c|}{Car Stop}&\multicolumn{2}{c|}{Color Cone}&\multicolumn{2}{c|}{Bump}&\multirow{2}{*}{mAcc}&\multirow{2}{*}{mIoU}\\ 
		\cline{2-15} 
		~& Acc& IoU& Acc& IoU & Acc & IoU& Acc& IoU& Acc& IoU& Acc& IoU& Acc& IoU& & \\
		\hline
		
		MST-SR&86.25 &82.36 &74.93 &65.40 &70.38 &60.21 &55.39 &40.21 &29.00 &26.28 &52.53 &\textcolor{blue}{47.67} &49.71 &44.53 &57.70 &51.65\\
		\hline
		DenseFuse&87.00 &81.46 &76.16 &68.25 &66.09 &57.90 &\textcolor{red}{57.65} &\textcolor{blue}{44.43} &15.20 &13.69 &36.73 &35.36 &39.72 &36.76 &53.10 &48.40\\
		\hline
		RFN-Nest&89.26 &83.49 &77.05 &67.98 &67.85 &59.68 &44.83 &35.23 &\textcolor{blue}{33.75} &\textcolor{blue}{28.47} &\textcolor{red}{54.10} &47.47 &46.28 &42.73 &56.94 &51.42\\
		\hline
		
		SMoA&89.28 &83.13 &78.30 &\textcolor{blue}{69.05} &66.44 &58.81 &48.89 &38.43 &23.96 &22.03 &49.31 &46.64 &47.69 &44.80 &55.92 &51.19\\
		\hline
		GANMcC&\textcolor{blue}{89.81} &81.46 &77.67 &67.82 &64.16 &57.36 &41.91 &33.59 &23.81 &20.85 &48.93 &45.15 &48.38 &41.72 &54.88 &49.51\\
		\hline
		U2Fusion&89.76 &\textcolor{blue}{83.86} &77.77 &68.63 &\textcolor{blue}{71.10} &\textcolor{blue}{61.62} &49.41 &36.76 &31.51 &28.19 &48.89 &46.00 &\textcolor{blue}{55.04} &\textcolor{red}{47.49} &\textcolor{blue}{58.08} &\textcolor{blue}{52.26}\\
		\hline
		MFEIF&88.30 &82.73 &76.23 &68.26 &63.61 &57.43 &51.48 &40.06 &27.08 &21.97 &44.85 &42.74 &45.90 &43.60 &55.20 &50.51\\
		\hline
		SeAFusion&89.43 &82.76 &\textcolor{blue}{78.42} &67.74 &66.07 &58.28 &52.31 &38.56 &28.58 &24.35 &46.73 &43.58 &47.10 &45.12 &56.43 &50.90\\
		\hline
		%		Ours&\textcolor{blue}{90.14} &\textcolor{blue}{84.24} &\textcolor{blue}{80.18} &\textcolor{blue}{71.06} &69.15 &60.81 &\textcolor{red}{59.33} &43.12 &32.02 &26.70 &52.71 &\textcolor{blue}{48.78} &\textcolor{red}{61.88} &\textcolor{red}{48.41} &\textcolor{blue}{60.51} &\textcolor{blue} {53.44}\\
		%		\hline
		Ours & \textcolor{red}{90.14} &\textcolor{red}{85.30} &\textcolor{red}{82.27} &\textcolor{red}{71.65} &\textcolor{red}{73.53} &\textcolor{red}{63.08} &\textcolor{blue}{56.24} &\textcolor{red}{44.51} &\textcolor{red}{35.92} &\textcolor{red}{30.96} &\textcolor{blue}{53.81} &\textcolor{red}{50.61} &\textcolor{red}{59.98} &\textcolor{blue}{47.38} &\textcolor{red}{61.24} &\textcolor{red} {54.61}\\
		\hline
	\end{tabular}
	\label{tab: SegmentationNum}
\end{table*}

\subsection{Loss Function}
	\subsubsection{Warm-start loss function}
	The warm-start fusion loss function can be formulated as: $\mathcal{L}_{WS}=\frac{1}{HW}\left \| I_{f}-\frac{I_{ir}+I_{vis}}{2} \right \| _{1},$
	where H and W denote the height and width of the source image, respectively. $I_{f}$ is the fused image. $\| \cdot \|_{1}$ represents the calculation of the $\ell_{1}$ norm.

	\subsubsection{Semantic training loss function}
	During the semantic training phase, we use the semantic training loss function as follows: $\mathcal{L}_{ST}=\mathcal{L}_{sem}+\lambda\mathcal{L}_{reg},$
	where the $\mathcal{L}_{sem}$ is a commonly used cross-entropy loss function, while the $\mathcal{L}_{reg}$ is a regularization term. The $\lambda$ is used to strike a balance between the two terms. We define the regularization loss function as:
	$
		\mathcal{L}_{reg}= \frac{1}{Corr\left(I_{ir},I_{f}\right) + Corr\left(I_{vis},I_{f}\right)},
	$
	where $Corr\left(\cdot\right)$ denotes the calculation of the correlation of two image tensors.
%	, which is a common statistical measure.
%	\begin{equation}
%		Corr\left(I_1,I_2\right)=\frac{Cov(I_1, I_2)}
%		{\sqrt{Var\left(I_1\right)Var\left(I_2\right)}},
%	\end{equation}
%	where the $Cov\left(\cdot,\cdot\right)$ represents the covariance between two image tensors. And the $Var\left(\cdot\right)$ denotes the variance of an image tensor. 

\section{Experiments}
In this section, we first conduct qualitative and quantitative  comparisons on three datasets: MFNet \cite{MFNet}, TNO \cite{TNO} and RoadScene \cite{U2Fusion}.
The competitors include MST-SR \cite{MST-SR}, DenseFuse \cite{DenseFuse}, RFN-Nest \cite{RFN-Nest}, SMoA \cite{SMoA}, GANMcC \cite{GANMcC}, U2Fusion \cite{U2Fusion}, MFEIF \cite{MFEIF}, and SeAFusion \cite{SeAFusion}.
We also conducted ablation studies to validate the proposed architecture and semantic-driven training strategy.

\subsection{Fusion Results}
%	\subsubsection{Qualitative comparisons}
	As shown in Fig.~\ref{fig: Qualitative}, we can see that our method can flexibly preserve abundant and useful textural details, successfully highlighting the important targets in diverse harsh environment. For instance, persons in our method have sharper edges, and thus stand out from the background, which will be helpful for follow-up segmentation.
\begin{figure}[htb]
	\centering
	\begin{tabular}{c}
		\includegraphics[width=0.47\textwidth]{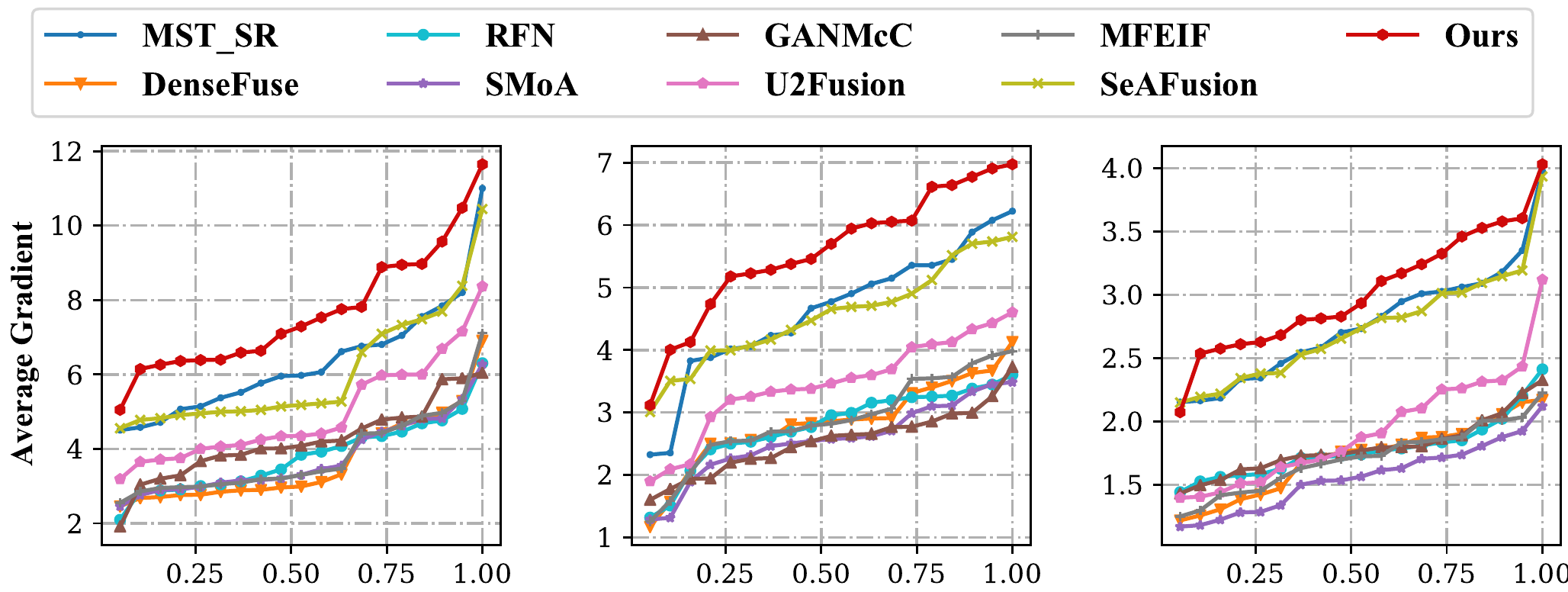}\\
		\includegraphics[width=0.47\textwidth]{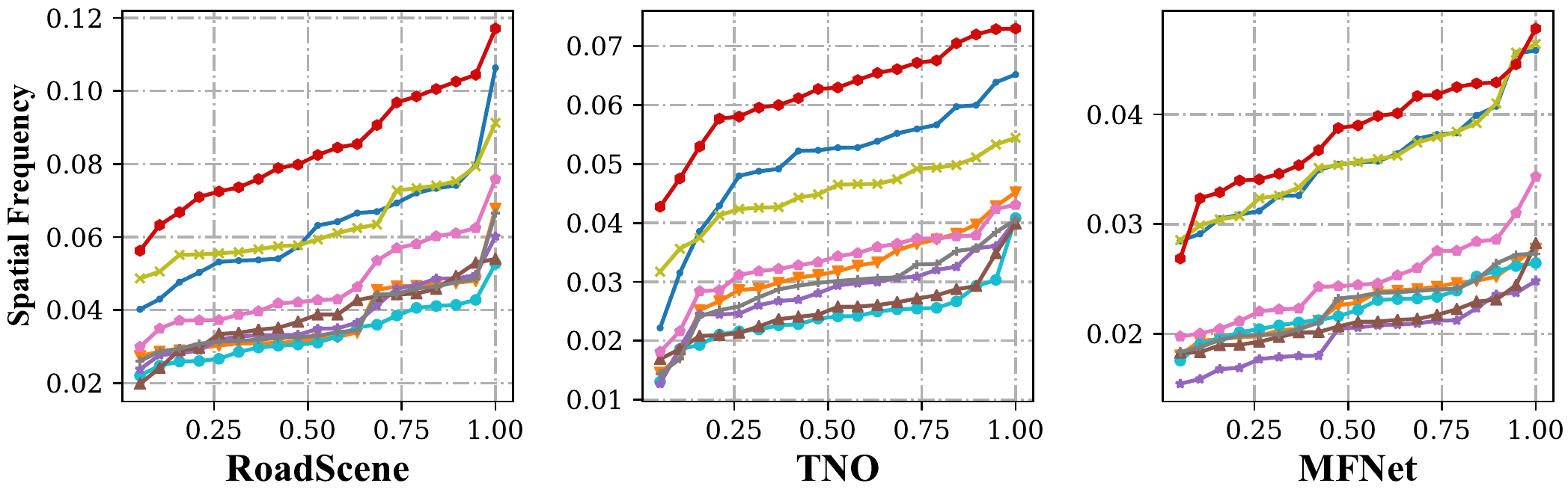}
	\end{tabular}
	\caption{Statistical evaluation of fusion results. A point $(x,y)$ denotes that there
		are $100 \ast x$ percent of image pairs which have values no more than y.}
	\label{classic metrics}
	
\end{figure}
%	\subsubsection{Quantitative Comparisons}
As for quantitative comparisons,
	Our method outperformed others in terms of commonly used statistical evaluation metrics spatial frequency (SF) \cite{SF} and average gradient (AG) \cite{AG}, as shown in Fig.~\ref{classic metrics}. This indicates our  results have more information richness, containing more  details and high contrast.
%\vspace{-1cm} 

\subsection{Segmentation results}
\begin{figure}[t]
	\setlength{\tabcolsep}{0.5pt}
	\centering
	\footnotesize
	\begin{tabular}{cccc}
		\includegraphics[width=0.12\textwidth]{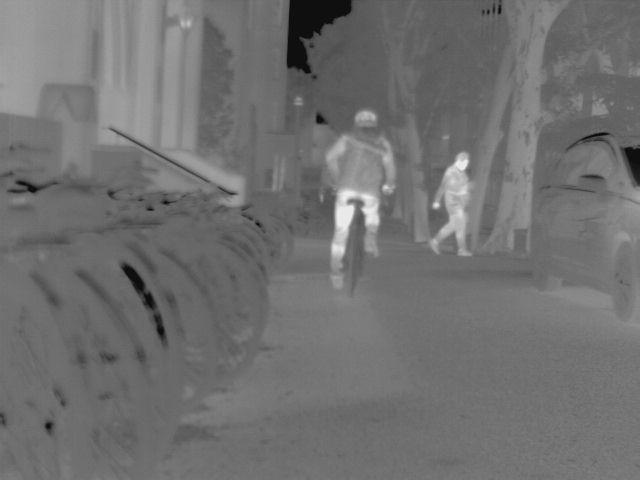}&
		\includegraphics[width=0.12\textwidth]{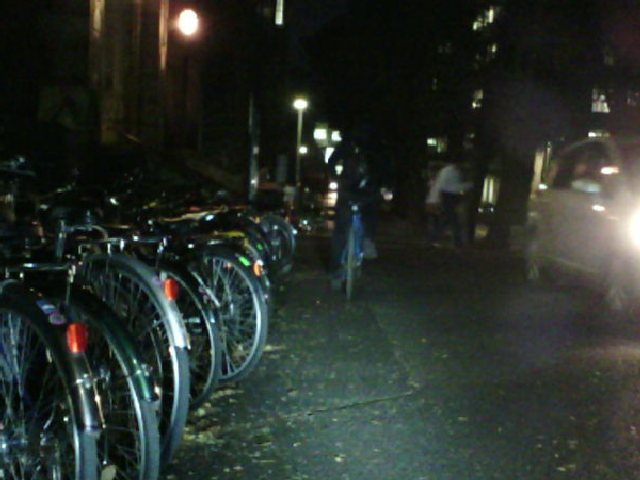}&
		\includegraphics[width=0.12\textwidth]{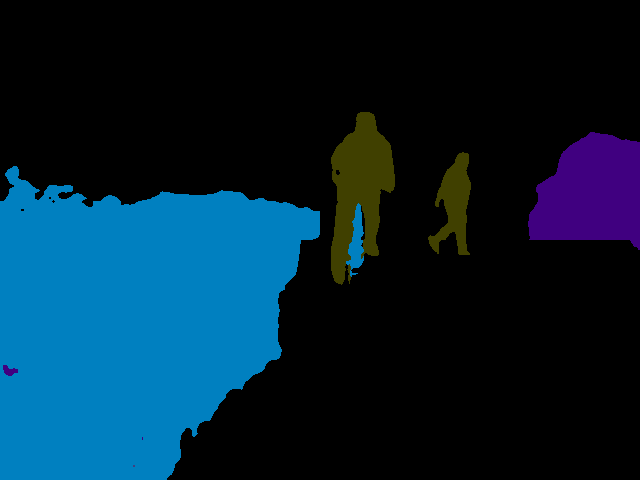}&
		\includegraphics[width=0.12\textwidth]{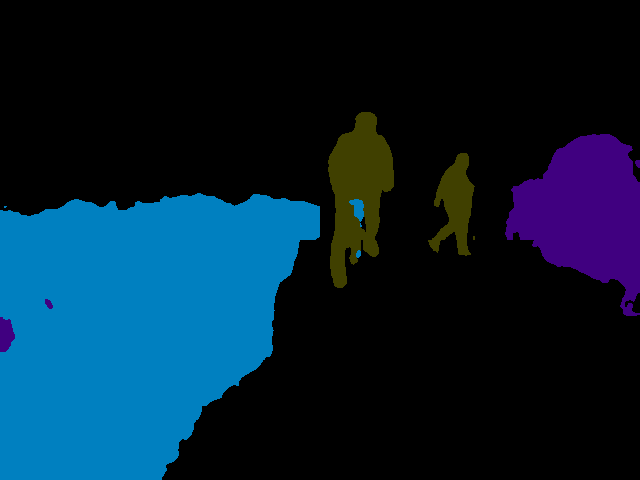}\\
		Ir&Vis&MST-SR&SMoA\\
		\includegraphics[width=0.12\textwidth]{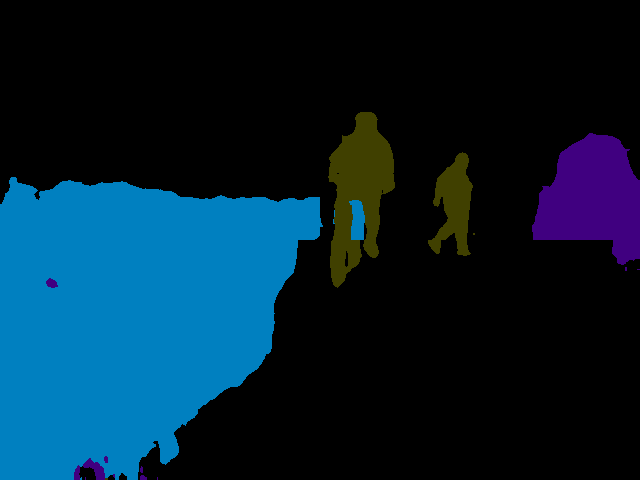}&
		\includegraphics[width=0.12\textwidth]{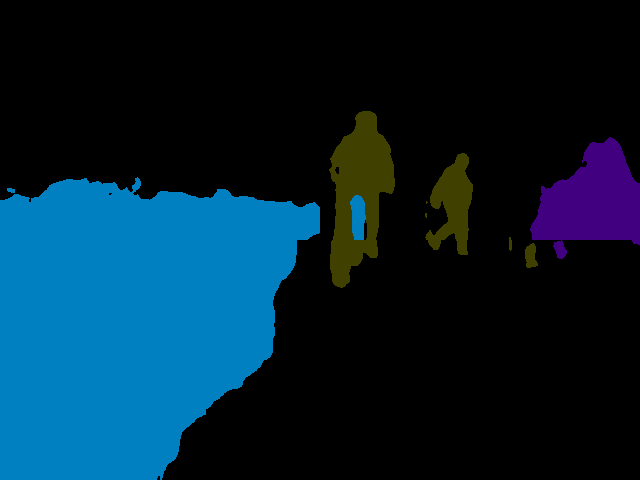}&
		\includegraphics[width=0.12\textwidth]{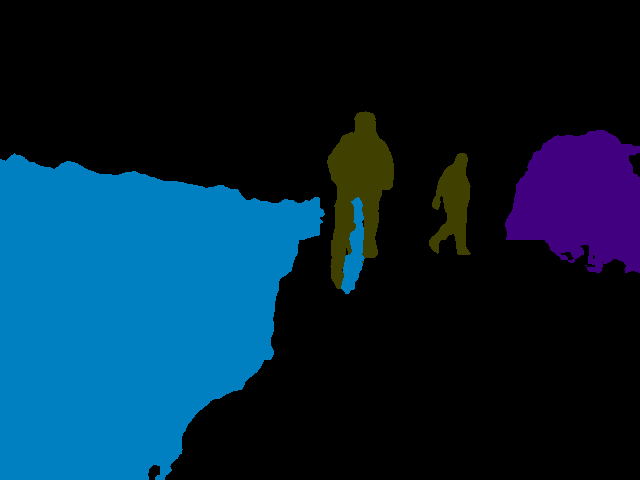}&
		\includegraphics[width=0.12\textwidth]{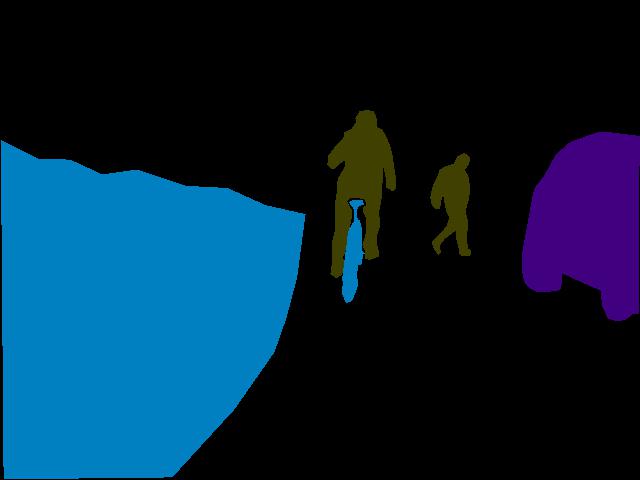}\\
		U2Fusion&RFN&Ours&Ground Truth
	\end{tabular}
	\caption{Qualitative comparison of segmentation results.}
	\label{Qualitative: seg}
\end{figure}

We also evaluate the fusion quality from the perspective of semantic segmentation. We trained the same segmentation network (SegFormer-b0) from scratch based on the fused images generated by all comparative methods. 
%	\subsubsection{Qualitative comparisons}
	 As shown in  Fig.~\ref{Qualitative: seg},  segmentation model with our method can  provide more accurate  results, e.g.,  bikes. In contrast, other methods cannot estimate the shapes of car, interfered by the strong glare.
%\vspace{-0.2cm} 
\begin{figure}[!htb]
	\centering
	\setlength{\tabcolsep}{0.5pt}
	\footnotesize
	
	%3、设置子图与上面正文或别的内容的距离
	\subfigtopskip=2pt
	%4、两行子图的间距
	\subfigbottomskip=2pt
	%5、子图和自己的label之间的距离
	\subfigcapskip=3pt
	
	\begin{tabular}{ccccc}
		
		\subfigure[Ir/Vis]{\label{net1}\includegraphics[width=0.095\textwidth]{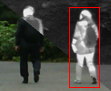}}&
		\subfigure[w/o SLA]{\label{net2}\includegraphics[width=0.095\textwidth]{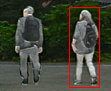}}&
		\subfigure[CHA]{\label{net3}\includegraphics[width=0.095\textwidth]{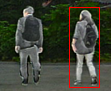}}&
		\subfigure[SPA]{\label{net4}\includegraphics[width=0.095\textwidth]{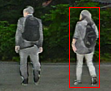}}&
		\subfigure[Ours]{\label{net5}\includegraphics[width=0.095\textwidth]{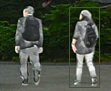}}
	\end{tabular}
	\caption{Visual effects using different network architectures.}
	\label{ablationnet}
\end{figure}
%\subsubsection{Quantitative comparisons}
%As shown in Table~\ref{tab: SegmentationNum}, our fusion results itself facilitate the segmentation task. $\mbox{Ours}^{+}$ achieves even better performance, implying that gradually interactive training is beneficial.
As shown in Table~\ref{tab: SegmentationNum}, our method got the top scores on mAcc and mIoU, indicating our method can intelligently reserve useful information for different semantic classes. Due to the information distortion of other methods, these methods cannot obtain the same accuracy of almost classes.
%	The Acc and IoU of the guardrail class are overwhelmingly zero, so we omit the results for the guardrail class in this table.

%\begin{table}[!htb]
%	%\small
%	\caption{Segmentation performance of different strategies. The best result is in {\textcolor{red}{\textbf{red}}} whereas the second best one is in {\textcolor{blue}{\textbf{blue}}}.}
%	\centering
%	\setlength\tabcolsep{7pt}
%	\renewcommand\arraystretch{1.1}
%	%\setlength\tabrowsep{5pt}
%	\footnotesize
%	% \resizebox{\textwidth}{12mm}{
%		\begin{tabular}{|c|c|c|c|c|c|}
%			\hline 
%			Structure&\multicolumn{2}{c|}{Spatial Frequency}&\multicolumn{2}{c|}{Average Gradient}&mIoU\\
%			\hline
%			w/o Attention &\multicolumn{2}{c|}{5.20}&\multicolumn{2}{c|}{5.20}&53.21\\
%			\hline
%			CHA &\multicolumn{2}{c|}{5.20}&\multicolumn{2}{c|}{5.20}&53.48 \\
%			\hline
%			SPA &\multicolumn{2}{c|}{5.20}&\multicolumn{2}{c|}{5.20}&\textcolor{blue}{53.78}  \\
%			\hline
%			Ours &\multicolumn{2}{c|}{5.20}&\multicolumn{2}{c|}{5.20}&\textcolor{blue}{53.78}  \\
%			\hline
%			\hline
%			Strategy&Car&Person&Car Stop&Bump&mIoU\\
%			\hline
%			Max-ST &\textcolor{red}{85.66}&\textcolor{red}{71.90}&23.71&47.91&53.70 \\
%			\hline
%			%			w/o WS\&Reg &83.70&71.39&15.39&41.44&50.3 \\
%			w/o WS &84.43&70.91&13.24&47.19&51.02 \\
%			\hline
%			w/o $\mathcal{L}_{reg}$ &85.05&71.28&\textcolor{blue}{25.86}&46.39&53.40 \\
%			\hline
%			Ours &\textcolor{blue}{85.30}&\textcolor{blue}{71.65}&\textcolor{red}{30.96}&\textcolor{red}{50.61}&\textcolor{red}{54.61} \\
%			\hline
%		\end{tabular}
%		\label{AblationNum}
%	\end{table}

\vspace{-0.4cm} 

%\begin{table}[htb]
%	\renewcommand{\arraystretch}{1.1}
%	\caption{{Qualitative comparison among different architecture-learning, parameter-learning strategies and  multi-scale fusion cell.}}
%	\label{tab:multi-task}
%	\centering
%	\setlength{\tabcolsep}{0.8mm}{
%		\begin{tabular}{|c|c|c|c|c|}
%			\hline
%			\multicolumn{2}{|c|}{Architecture Learning} & Parameter Learning &{Denoising}&{Deraining}\\
%			\hline
%			\multirow{5}{*}{{Darts}}&$\mathtt{N}$&\multirow{5}{*}{Single-task Learning}&36.94/0.965&36.60/0.977\\
%			&$0.7\mathtt{N}+0.3\mathtt{R}$&&36.95/0.964&37.21/0.980\\   
%			&$0.5\mathtt{N}+0.5\mathtt{R}$&&36.81/0.962&36.93/0.978\\
%			&$0.3\mathtt{N}+0.7\mathtt{R}$&&36.68/0.963&36.82/0.978\\
%			&$\mathtt{R}$&&36.52/0.953&37.31/0.981\\
%			\hline
%			\multirow{2}{*}{{Cross-Darts}}&\multirow{2}{*}{$0.5\mathtt{N}+0.5\mathtt{R}$}&Multi-task Learning&34.11/0.887&33.17/0.943\\
%			\cline{3-5}
%			~&~&Single-task Learning&\textbf{36.98/0.965}&\textbf{37.54/0.981}\\\hline\hline
%			Nums of Scale &1& 2 & 3 & 4\\
%			\hline
%			PSNR/SSIM  & 34.69/0.971 &37.90/0.983&38.27/0.983& 38.19/0.985\\
%			\hline
%	\end{tabular}}
%\end{table}

\subsection{Ablation studies}

	\subsubsection{Analyzing the fusion module}
	We conducted experiments including removing the self-attention module (w/o SLA), replacing it with channel attention (CHA)  \cite{SEN}, and spatial attention (SPA)  \cite{SCA}.
	As shown in Fig.~\ref{ablationnet},  persons of our fused result are more natural and conspicuous, which is also demonstrated by segmentation results, as reported in Table~\ref{AblationNum}.
%\vspace{-0.2cm} 

%\vspace{-0.6cm} 
\begin{table}[!htb]
	%\small
	\caption{Segmentation results of different structures and strategies.}
	\centering
	\setlength\tabcolsep{5.8pt}
	\renewcommand\arraystretch{1.1}
	\footnotesize
	% \resizebox{\textwidth}{12mm}{
		\begin{tabular}{|c|c|c|c|c|c|c|c|}
			\hline 
			\multicolumn{2}{|c|}{}&Car&Person&Car Stop&Bump&mIoU\\
%			\multicolumn{2}{|c|}{\;}&Car&Person&Car Stop&Bump&mIoU\\
			\hline
			\multirow{3}*{Structure}&w/o SLA &85.11 &71.25&21.55&48.96&53.21\\
			%			\hline
			\cline{2-7}
			&CHA &84.37&71.53&23.86&\textcolor{blue}{49.10}&53.48 \\
			\cline{2-7}
			%			\hline
			&SPA &84.90&71.30&24.31&48.11&\textcolor{blue}{53.78}  \\

			\hline
			\hline
			\multirow{3}*{Strategy}&Max-ST &\textcolor{red}{85.66}&\textcolor{red}{71.90}&23.71&47.91&53.70 \\
			\cline{2-7}
			%			\hline
			%			w/o WS\&Reg &83.70&71.39&15.39&41.44&50.3 \\
			&w/o WS &84.43&70.91&13.24&47.19&51.02 \\
			%			\hline
			\cline{2-7}
			&w/o $\mathcal{L}_{reg}$ &85.05&71.28&\textcolor{blue}{25.86}&46.39&53.40 \\
			\cline{2-7}
			\hline
			\hline
			
			\multicolumn{2}{|c|}{Ours (Ave-ST)} &\textcolor{blue}{85.30}&\textcolor{blue}{71.65}&\textcolor{red}{30.96}&\textcolor{red}{50.61}&\textcolor{red}{54.61} \\
			\hline
		\end{tabular}
		\label{AblationNum}
	\end{table}

\subsubsection{Analyzing the warm-start and regularization loss}
\begin{figure}[htb]
	\centering
	\setlength{\tabcolsep}{0.5pt}
	\footnotesize
	
	%3、设置子图与上面正文或别的内容的距离
	\subfigtopskip=-1pt
	%4、两行子图的间距
	\subfigbottomskip=2pt
	%5、子图和自己的label之间的距离
	\subfigcapskip=3pt
	\begin{tabular}{ccccc}
		
		\subfigure[Ir/Vis]{\label{classA}\includegraphics[width=0.095\textwidth]{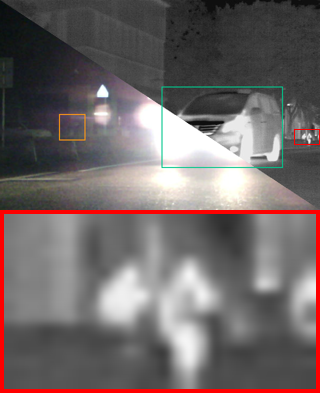}}&
		%		\subfigure[w/o WS\&Reg]{\label{classB}\includegraphics[width=0.12\textwidth]{results/ablation/ws/naive.png}}&
		%		\subfigure[w/o WS]{\label{classC}\includegraphics[width=0.12\textwidth]{results/ablation/ws/naive_reg.png}}&
		%		\subfigure[w/o Reg]{\label{classD}\includegraphics[width=0.12\textwidth]{results/ablation/ws/noreg.png}}
		
		%		\\
		\subfigure[Max]{\label{classE}\includegraphics[width=0.095\textwidth]{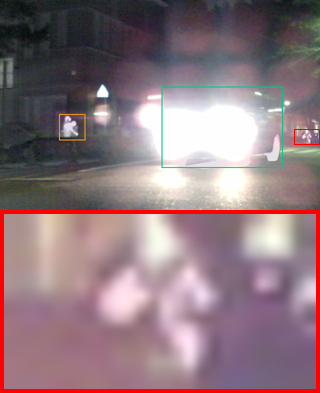}}&
		\subfigure[Ave]{\label{classF}\includegraphics[width=0.095\textwidth]{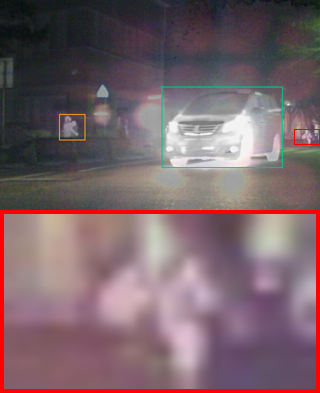}}&
		\subfigure[Max-ST]{\label{classG}\includegraphics[width=0.095\textwidth]{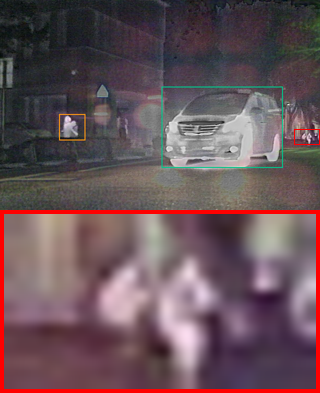}}&
		\subfigure[Ave-ST]{\label{classH}\includegraphics[width=0.095\textwidth]{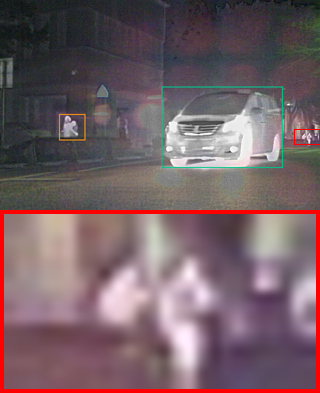}}
	\end{tabular}
	\caption{Intermediate visual effects and  subsequent results of  warm-start.}
	\label{ablationws}
\end{figure}

 We also provide another warm start, i.e., $\mathcal{L}_{WS}=\frac{1}{HW}\left \| I_{f}-max\left(I_{ir}, I_{vis}\right) \right \| _{1},$ where $max\left(\cdot\right)$ is denoted as element-wise maximum selection. The fused results after the warm-start  and semantic training  are denoted as ``Max'' and ``Max-ST'', respectively. The corresponding results of proposed scheme are denoted as ``Ave'' and ``Ave-ST''. 
As shown in Fig.~\ref{ablationws}, (b) fails to suppress glare and (c) fails to emphasize pedestrians. After the semantic training phase, they overcome these problems (i.e., (d) and (e)). As shown in Table~\ref{AblationNum},  maximum selection fusion rule provides better initialization to preserve the pedestrians and cars. But it fails to deal with other classes, such as car stop. This shows that average fusion rule can provide a more malleable initialization.
In addition, we conducted experiments including removing the warm-start phase (w/o WS), and removing the regularization term (w/o $\mathcal{L}_{reg}$). As shown in Table~\ref{AblationNum},  segmentation results are decreased without these strategies.
%\vspace{-0.3cm} 

	\subsubsection{Analyzing the effect of semantic loss}
\begin{figure}[htb]
	\centering
	\setlength{\tabcolsep}{0.5pt}
	\footnotesize
	%3、设置子图与上面正文或别的内容的距离
	\subfigtopskip=-5pt
	%4、两行子图的间距
	\subfigbottomskip=2pt
	%5、子图和自己的label之间的距离
	\subfigcapskip=3pt
	\begin{tabular}{ccc}
		
		\subfigure[Ir/Vis]{\label{class1}\includegraphics[width=0.16\textwidth]{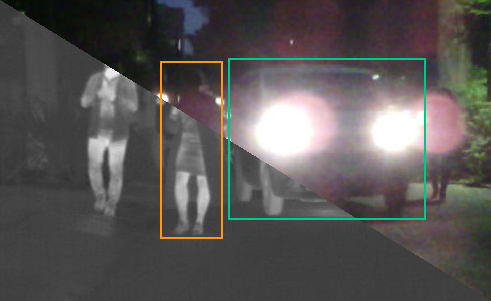}}&
		\subfigure[w/o $\mathcal{L}_{sem}$]{\label{class2}\includegraphics[width=0.16\textwidth]{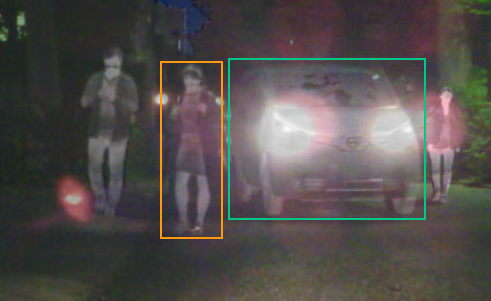}}&
		\subfigure[Ours]{\label{class3}\includegraphics[width=0.16\textwidth]{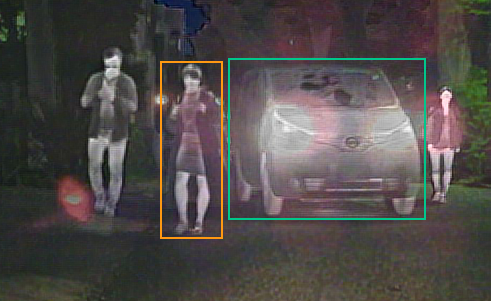}}\\
		
		\\
		\subfigure[w/o Car\&Person]{\label{class4}\includegraphics[width=0.16\textwidth]{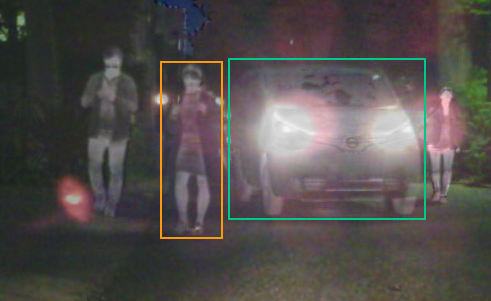}}&
		\subfigure[w/o Car]{\label{class5}\includegraphics[width=0.16\textwidth]{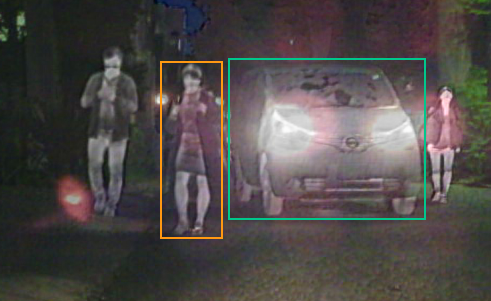}}&
		\subfigure[w/o Person]{\label{class6}\includegraphics[width=0.16\textwidth]{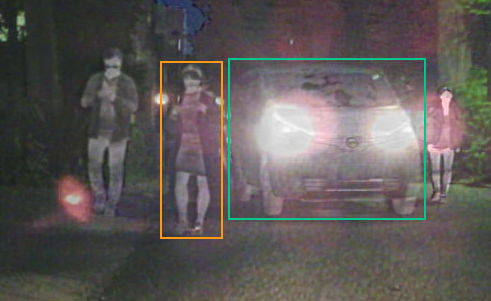}}\\
	\end{tabular}
	\caption{Visual comparison by removing different semantic classes.}
	\label{ablationclass}
	
\end{figure}

We conducted experiments removing some classes, e.g., the car class (w/o Car), the person class (w/o Person), and the both (w/o Car\&Person). We also conducted an experiment removing the semantic loss (w/o $\mathcal{L}_{sem}$). As shown in Fig.~\ref{ablationclass}, the semantic loss makes the pedestrians and the cars stand out from the background as shown in (c). The impact of person class is more significant than the car class. This is because in the MFNet, pedestrians has a larger number than car. It is worth  to finding that (b) is totally the same as (d), implying that the infrared information of other classes than pedestrian and car classes contribute less to the segmentation task.
%\vspace{-0.3cm} 
\section{Conclusion}

In this letter, to break free from manually designing fusion rules, we develop a semantic-level fusion network with an adaptive semantic-driven training strategy to take full advantage of the guidance from the follow-up semantic tasks. Experimental results reveal that semantic loss can not only replace the manual design of fusion rules but also provide a flexible and robust semantic-level fusion that satisfies both human vision and high-level vision tasks.

\clearpage
\balance
\bibliographystyle{IEEEtran}
\bibliography{BFFR}

\end{document}